\documentclass[sigconf]{acmart}
\usepackage{graphicx}
\usepackage{amsmath,amsfonts}
\usepackage{amsthm}
\usepackage{booktabs}
\usepackage{multirow}
\usepackage{bm}
\usepackage{subfigure}
\usepackage{xcolor}
\usepackage[linesnumbered,ruled,noend]{algorithm2e} 
\usepackage{algorithmic}
\usepackage{siunitx}
\usepackage[switch]{lineno}
\usepackage{paralist}
\usepackage{chngcntr}
\usepackage{wrapfig}
\usepackage{adjustbox}
\usepackage{siunitx}
\usepackage{paralist}
\AtBeginDocument{%
  \providecommand\BibTeX{{%
    \normalfont B\kern-0.5em{\scshape i\kern-0.25em b}\kern-0.8em\TeX}}}

\copyrightyear{2021}
\acmYear{2021}
\setcopyright{acmcopyright}
\acmConference[KDD '21]{Proceedings of the 27th ACM
SIGKDD Conference on Knowledge Discovery and Data Mining}{August 14--18, 2021}{Virtual Event, Singapore}
\acmBooktitle{Proceedings of the 27th ACM SIGKDD Conference on Knowledge Discovery and Data Mining (KDD '21), August 14--18, 2021, Virtual Event, Singapore}
\acmPrice{15.00}
\acmDOI{10.1145/3447548.3467419}
\acmISBN{978-1-4503-8332-5/21/08}

\begin{document}
\fancyhead{}
\title{ACE-NODE: Attentive Co-Evolving Neural Ordinary Differential Equations}

\author{Sheo Yon Jhin, Minju Jo, Taeyong Kong, Jinsung Jeon, and Noseong Park}
\email{{sheoyonj, alflsowl12, qbxlvnf11, jjsjjs0902, noseong}@yonsei.ac.kr}
\affiliation{\institution{Yonsei University}
  \city{Seoul}
  \country{South Korea}}

\renewcommand{\shortauthors}{Trovato and Tobin, et al.}

\begin{abstract}
Neural ordinary differential equations (NODEs) presented a new paradigm to construct (continuous-time) neural networks. While showing several good characteristics in terms of the number of parameters and the flexibility in constructing neural networks, they also have a couple of well-known limitations: i) theoretically NODEs learn homeomorphic mapping functions only, and ii) sometimes NODEs show numerical instability in solving integral problems. To handle this, many enhancements have been proposed. To our knowledge, however, integrating attention into NODEs has been overlooked for a while. To this end, we present a novel method of attentive dual co-evolving NODE (ACE-NODE): one main NODE for a downstream machine learning task and the other for providing attention to the main NODE. Our ACE-NODE supports both pairwise and elementwise attention. In our experiments, our method outperforms existing NODE-based and non-NODE-based baselines in almost all cases by non-trivial margins.
\end{abstract}





\maketitle

\section{Introduction}

It is known that the residual connection, denoted $\bm{h}(t+1) = \bm{h}(t) + f(\bm{h}(t))$ where $\bm{h}(t)$ means a hidden vector at layer (or time) $t$, is identical to the explicit Euler method to solve ODEs~\cite{NIPS2018_7892}. In this regard, neural ordinary differential equations (NODEs) are to generalize the residual connection with a continuous time variable $t$. In other words, $t$ in NODEs can be an arbitrary real number whereas it must be a non-negative integer in conventional residual networks. NODEs show their efficacy in many different tasks~\cite{2019arXiv191010470P,yan2020robustness,zhuang2020ordinary}. They sometimes not only show better accuracy but also have a smaller number of parameters in comparison with conventional neural networks.

In order to further improve, many researchers proposed enhancements for NODEs, ranging from ODE state augmentation to novel regularizations specific to NODEs~\cite{NIPS2019_8577,finlay2020train}. The ODE state augmentation was proposed to overcome the homeomorphic characteristic of NODEs and various regularization concepts were proposed to learn straight-line ODEs that are considered to be easy to solve.

\begin{figure}[t]
\centering
\subfigure[The original NODE architecture]{\includegraphics[width=1\columnwidth]{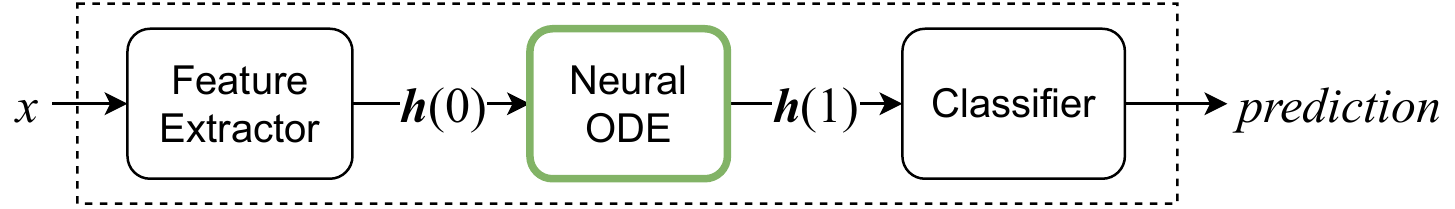}}
\subfigure[Our proposed ACE-NODE architecture]{\includegraphics[width=1\columnwidth]{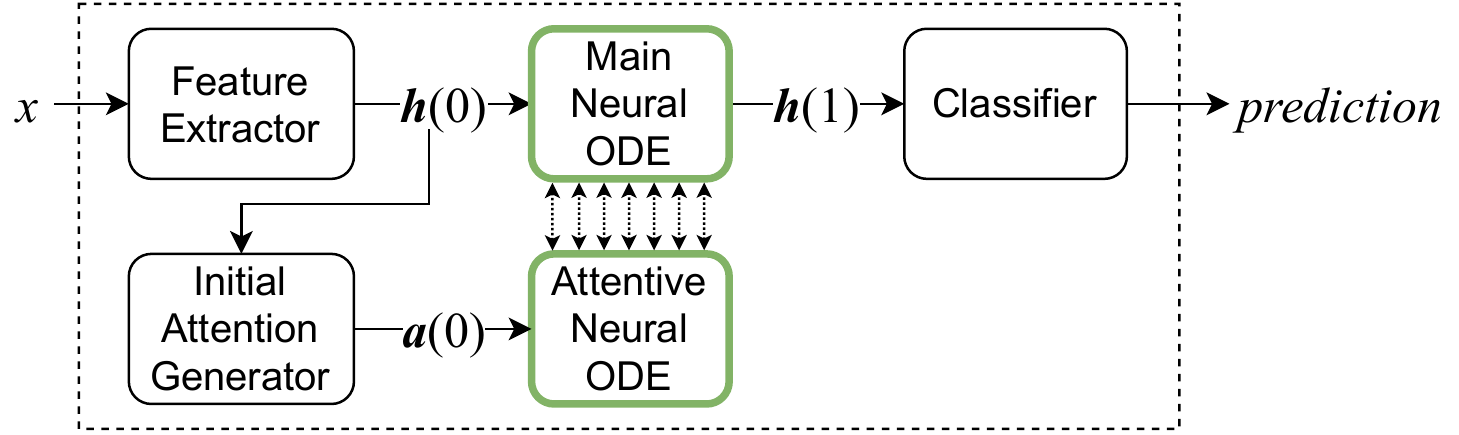}}
\caption{The general architecture of NODEs and our proposed ACE-NODEs. In our design, the main NODE and the attentive NODE co-evolve over time and influence each other.}    \label{fig:archi}
\end{figure}

To our knowledge, however, there is no existing work to integrate the concept of attention into NODEs. To this end, we design a novel model of \textbf{A}ttentive dual \textbf{C}o-\textbf{E}volving NODE (ACE-NODE) and its training algorithm: one NODE is for describing a time-evolving process of hidden vector, denoted $\bm{h}(t)$ in our paper, and the other is for a time-evolving process of attention, denoted $\bm{a}(t)$ (cf. Fig.~\ref{fig:archi} (b)). In our proposed model, therefore, $\bm{h}(t)$ and $\bm{a}(t)$ co-evolve to accomplish a downstream machine learning task, e.g., image classification, time-series forecasting, and so forth.

As shown in Fig.~\ref{fig:archi}, our design follows the original NODE architecture which includes feature extractor and classifier --- for simplicity but without loss of generality, we assume classification in this architecture. After generating the initial attention, denoted $\bm{a}(0)$ in the figure, however, we have dual co-evolving NODEs, which makes our model more sophisticated than the original design, i.e., the original architecture in Fig.~\ref{fig:archi} (a) vs. our architecture in Fig.~\ref{fig:archi} (b), and as a result, its training algorithm also becomes more complicated.

In our work, the attention produced by $\bm{a}(t)$ can be categorized into two types: i) pairwise attention and ii) elementwise attention --- there also exist other types of attention~\cite{graves2014neural,luong-etal-2015-effective,DBLP:conf/iclr/BahdanauHLHKG18}, which frequently happens in natural language processing and we leave them as future work because our main target is not natural language processing in this paper. The former frequently happens in time-series forecasting and the latter in image classification. The pairwise attention learns an attention value from $i$-th to $j$-th dimension of $\bm{h}(t)$ for all $i,j$. In the case of multivariate time-series forecasting, each dimension of $\bm{h}(t)$ means a time-series value that is potentially correlated with other time-series values (dimensions), e.g., the first dimension of $\bm{h}(t)$ is the stock price of Google, the second dimension is that of Apple, and so forth. Therefore, it is natural to learn attention among them because those time-series values are correlated in real-world environments. The proposed ACE-NODE is rather complicated than other single NODE applications. Therefore, we propose the formal definition of our proposed concept, followed by its gradient computation and training methods.

In the case of image classification, $\bm{h}(t)$ means a continuous-time feature map. For instance, ODE-Net, which follows the architecture in Fig.~\ref{fig:archi} (a), uses image convolution operations to define its NODE layer~\cite{NIPS2018_7892}. Therefore, $\bm{h}(t)$ means a convolutional feature map for all $t$. In such a case, our attentive NODE $\bm{a}(t)$ produces an elementwise attention, which will be elementwise multiplied with $\bm{h}(t)$. One can consider that our attentive NODE learns which information in the feature map is important.

We conduct experiments with time-series and image datasets. For image classification, we use MNIST~\cite{lecun2010mnist}, SVHN~\cite{Netzer2011}, and CIFAR10~\cite{Krizhevsky09learningmultiple} and compare our method with ODE-Net and some other conventional convolutional networks. For time-series forecasting, we use a climate dataset and a hospital dataset. In almost all experiments, our methods show non-trivial enhancements over baselines by adding attention. Our contributions are summarized as follows:
\begin{enumerate}
    \item We define dual co-evolving NODEs, one is for learning hidden vectors and the other is for learning attention.
    \item We design a training mechanism specialized to the proposed dual co-evolving NODEs.
    \item We design its training method and show that our formulation is theoretically well-posed, i.e, its solution always exists and is unique, under the mild assumption of analytic ODE functions.
    \item We conduct experiments for various downstream tasks and our method shows the best accuracy in all seven experiments.
\end{enumerate}

\section{Related Work}
In this section, we review NODEs and attention. We also mention about our attention design goal in NODEs.

\subsection{Neural ODEs}\label{sec:node}
NODEs solve the following integral problem to calculate $\bm{h}(t_1)$ from $\bm{h}(t_0)$~\cite{NIPS2018_7892}:
\begin{linenomath*}\begin{align}
    \bm{h}(t_1) = \bm{h}(t_0) + \int_{t_0}^{t_1}f(\bm{h}(t),t;\bm{\theta}_f)dt,
\end{align}\end{linenomath*}where $f(\bm{h}(t),t;\bm{\theta}_f)$, which we call \emph{ODE function}, is a neural network to approximate $\dot{\bm{h}} \stackrel{\text{def}}{=} \frac{d \bm{h}(t)}{d t}$. To solve the integral problem, NODEs rely on ODE solvers, such as the explicit Euler method, the Dormand--Prince (DOPRI) method, and so forth~\cite{DORMAND198019}.

In general, ODE solvers discretize time variable $t$ and convert an integral into many steps of additions. For instance, the explicit Euler method can be written as follows in a step:
\begin{linenomath*}\begin{align}\label{eq:euler}
\bm{h}(t + s) = \bm{h}(t) + s \cdot f(\bm{h}(t), t;\bm{\theta}_f),
\end{align}\end{linenomath*}where $s$, which is usually smaller than 1, is a pre-determined step size of the Euler method. Note that this equation is identical to a residual connection when $s=1$. The DOPRI method uses a much more sophisticated method to update $\bm{h}(t + s)$ from $\bm{h}(t)$ and dynamically controls the step size $s$. However, those ODE solvers sometimes incur unexpected numerical instability~\cite{zhuang2020adaptive}. For instance, the DOPRI method sometimes keeps reducing the step size $s$ and eventually, an underflow error is produced. To prevent such unexpected problems, several countermeasures were also proposed.

One distinguished characteristic of NODEs is that we can calculate the gradient of loss w.r.t. NODE parameters, denoted $\nabla_{\bm{\theta}_f} L = \frac{d L}{d \bm{\theta}_f}$ where $L$ is a task-dependent loss function, can be calculated by a reverse-mode integration, whose space complexity is $\mathcal{O}(1)$ and more efficient than backpropagation. This method is called the adjoint sensitivity method.

Fig.~\ref{fig:archi} (a) shows the typical architecture of NODEs --- we assume a downstream classification task in this architecture. There is a feature extraction layer which provides $\bm{h}(t_0)$, where $t_0 = 0$, and $\bm{h}(t_1)$, where $t_1 = 1$, is calculated by the method described above. The following classification layer outputs a prediction for input $x$.

\begin{figure}
    \centering
    \includegraphics[width=0.45\columnwidth]{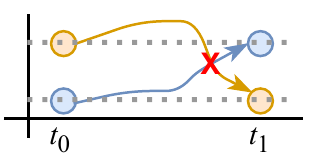}
    \caption{The locations of the yellow and blue points are swapped after the mapping from $t_0$ to $t_1$. The yellow and blue trajectories, which cross each other, cannot be learned by NODEs because their topology (i.e., their relative positions) cannot be changed after a homeomorphic mapping.}
    \label{fig:cross}
\end{figure}

\paragraph{Homeomorphic characteristic of NODEs.} Let $\phi_t : \mathbb{R}^{\dim(\bm{h}(t_0))} \rightarrow \mathbb{R}^{\dim(\bm{h}(t_1))}$ be a mapping from $t_0$ to $t_1$ created by a NODE after solving the integral problem. It is well-known that $\phi_t$ becomes a homeomorphic mapping: $\phi_t$ is continuous and bijective and $\phi_t^{-1}$ is also continuous for all $t \in [0,T]$, where $T$ is the last time point of the time domain~\cite{NIPS2019_8577,massaroli2020dissecting}. From this characteristic, the following proposition can be derived:
\begin{proposition}
The topology of the input space of $\phi_t$ is preserved in its output space, and therefore, trajectories crossing each other cannot be represented by NODEs, e.g., Fig.~\ref{fig:cross}.
\end{proposition}

While preserving the topology, NODEs can perform machine learning tasks and it was shown in~\cite{yan2020robustness} that it increases the robustness of representation learning to adversarial attacks. At the same time, however, Dupont et al. showed that NODEs' representation learning capability is sometimes not as high as expectation by the same reason and proposed a method to overcome the homeomorphic mapping limitation of NODEs~\cite{NIPS2019_8577}, which is called \emph{augmented NODE}. The augmentation method brings $\bm{h}(t_0)$ into a higher dimensional space by appending several zeros, i.e., $\bm{h}(t_0) \oplus \bm{0}$ where $\oplus$ means a concetenation and $\bm{0}$ is a zero vector in a certain dimensionality. They showed that $\dim(\bm{0}) = 5$ works well in most cases.

We consider that our attentive co-evolving NODE concept is also an effective way to increase the representation learning capability of NODEs. In our experiments, we conduct in-depth comparisons with various NODE models, including the augmentation-based NODE models.
 
\subsection{Attention}
Attention, which is to catch useful information, is one of the most successful concepts of deep learning. It had been early studied for natural language processing and computer vision~\cite{DBLP:journals/corr/BahdanauCB14,7780872,10.5555/3295222.3295349, spratling2004attention} and quickly spread to other fields~\cite{7243334,kim2018attentive,9194070}.

The attention mechanism can be best described by the human visual perception system, which focuses on selective parts of input while ignoring irrelevant parts~\cite{pmlr-v37-xuc15}. There exist several different types of attention and their applications. For machine translation, a soft attention mechanism was proposed by Bahdanau et al.~\cite{DBLP:journals/corr/BahdanauCB14}. A self-attention method was proposed to be used in language understanding~\cite{shen2018disan}. Kiela et al. proposed a general purpose text representation learning method with attention in~\cite{kiela2018dynamic}. A co-attention method was used for visual question answering~\cite{10.5555/3157096.3157129}. We refer to survey papers~\cite{DBLP:journals/corr/abs-1904-02874,10.1145/3363574} for more detailed information. Likewise, there exist many different types and applications of attention.

To our knowledge, however, it had not been actively studied yet for NODEs because NODEs are a relatively new paradigm of designing neural networks and it is not straightforward how to integrate attention into them. ETN-ODE~\cite{gao2020explainable} uses attention in its feature extraction layer before the NODE layer. It does not propose any new NODE model that is \emph{internally} combined with attention. In other words, they use attention to derive $\bm{h}(0)$ in their feature extraction layer, and then use a standard NODE layer to evolve $\bm{h}(0)$. It is the feature extraction layer which has attention in ETN-ODE. In this regard, it is hard to say that ETN-NODE is an attention-based NODE model.

In NODEs, however, $\bm{h}(t)$ means a hidden vector at time $t$. To help a downstream task with attention, we need to define $\bm{a}(t)$ which means attention at time $t$. After that, $\bm{h}(t + s)$ should be derived from $\bm{h}(t)$ aided by $\bm{a}(t)$. One more thing is that they naturally co-evolve rather than being independent from each other. We propose ACE-NODE in this notion in this paper.

\section{Attentive Neural ODEs}


In this section, we first introduce our proposed attention definition and then describe our co-evolving dual NODE training method. Our general architecture is in Fig.~\ref{fig:archi} (b). There is an initial attention generation function $q$, i.e., $\bm{a}(0)  = q (\bm{h}(0); \bm{\theta}_q)$. After that, our proposed dual co-evolving NODEs begin. $\bm{h}(t)$ and $\bm{a}(t)$ evolve over time together in our framework and they influence each other to produce a reliable hidden representation $\bm{h}(1)$ for an input sample $x$. There exists a classifier which will process $\bm{h}(1)$ and produce a prediction for $x$.

\subsection{Pairwise Attention}
Our proposed pairwise attention definition can be described by the following co-evolving dual NODEs:
\begin{linenomath*}\begin{align}\begin{split}\label{eq:panode}
    \bm{h}(t_1) =& \bm{h}(t_0) + \int_{t_0}^{t_1}f(\bm{h}(t),\bm{a}(t),t;\bm{\theta}_f)dt,\\
    \bm{a}(t_1) =& \bm{a}(t_0) + \int_{t_0}^{t_1}g(\bm{a}(t),\bm{h}(t),t;\bm{\theta}_g)dt,
\end{split}\end{align}\end{linenomath*}where $\bm{a}(t) = \mathbb{R}^{dim(\bm{h}(t)) \times dim(\bm{h}(t))}$. In other words, $\bm{a}(t)$ is a time-evolving matrix denoting the logit value of attention from $i$-th to $j$-th dimension of $\bm{h}(t)$ for all $i$ and $j$ pairs. As will be described shortly, we use a softmax activation to convert the logit into attention (cf. Eq.~\eqref{eq:aggr1}).

\begin{figure}[t]
\centering
\subfigure[]{\includegraphics[width=0.45\columnwidth]{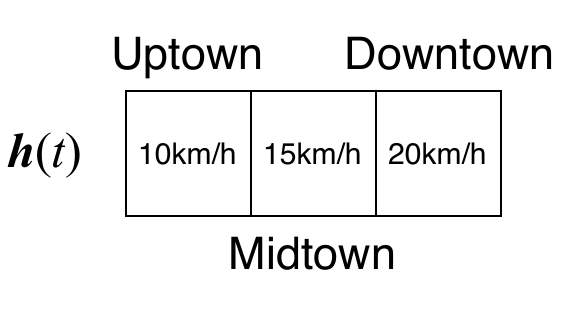}}\hspace{2em}
\subfigure[]{\includegraphics[width=0.25\columnwidth]{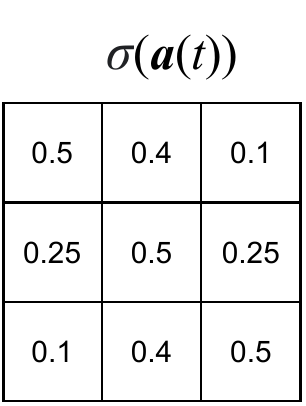}}
\caption{ODE state $\bm{h}(t)$ and pairwise attention $\sigma(\bm{a}(t))$ examples. (a) A 3-dimensional ODE state describing the traffic condition of the three regions of New York (b) A $3 \times 3$ pairwise attention matrix describing the attention among the three regions. Each region has the highest attention to itself, followed by attention to neighboring regions. This attention says that a traffic condition in a region at time $t + s$ has high correlations to the traffic conditions in the same and neighboring regions at time $t$.}    \label{fig:ny}
\end{figure}

This setting frequently happens for multivariate time-series forecasting, where $\bm{h}(t)$ means a snapshot of multivariate time-series at time $t$. In Fig.~\ref{fig:ny}, for instance, we show an ODE example describing the traffic condition of the three regions of New York. At time $t$, each element of $\bm{h}(t)$ contains average vehicle speed for each region. We also show an attention matrix example in the figure, where each region has strong attention to itself and its neighboring regions.

Therefore, we propose to apply the attention to $\bm{h}(t)$ at the beginning of $f$ as follows:
\begin{linenomath*}\begin{align}\begin{split}\label{eq:aggr1}
f(\bm{h}(t),\bm{a}(t),t;\bm{\theta}_f) &= f'(\bm{h}'(t),t;\bm{\theta}_f),\\
\bm{h}'(t) &= \bm{h}(t) \sigma(\bm{a}(t))^\intercal,
\end{split}\end{align}\end{linenomath*}where $\sigma$ is a softmax activation, and ``$\intercal$'' means transpose. We redefine the ODE function $f(\bm{h}(t),\bm{a}(t),t;\bm{\theta}_f)$ as $f'(\bm{h}'(t),t;\bm{\theta}_f)$ based on $\bm{h}'(t)$. We note that $dim(\bm{h}'(t)) = dim(\bm{h}(t))$. The other ODE function $g$ can also be defined as $g'(\bm{h}'(t),t;\bm{\theta}_g)$.

The definitions of $f'$ and $g'$ vary from one dataset/task to another. Therefore, we introduce our specific designs for them in our experimental evaluation section after describing a general form in this section. However, we can adopt a well-defined ODE function from other work for $f'$. For instance, GRU-ODE is a continuous generalization of gated recurrent units (GRUs)~\cite{debrouwer2019gruodebayes}. The authors of GRU-ODE proved that GRUs can be modeled theoretically by time-dependent ODEs. Therefore, we can adopt their ODE function, which models continuous-time GRUs, as our ODE function $f'$ and it becomes an attentive GRU-ODE with our additional attentive NODE.

In addition, we need to design one more ODE function $g'$. However, we found that each task or each data has its own optimized attention mechanism. Therefore, we introduce our specific design of $g'$ for each task or each data in the experimental evaluation section. We describe only our general design in this section.

\subsection{Elementwise Attention}
Our proposed elementwise attention definition can be described by Eq.~\eqref{eq:panode} with a different definition of $\bm{a}(t) = \mathbb{R}^{dim(\bm{h}(t))}$. This setting frequently happens in convolutional neural networks, where $\bm{h}(t)$ means a feature map created by convolutions.

We propose to perform the following elementwise multiplication to apply the attention to the feature map:
\begin{linenomath*}\begin{align}\begin{split}\label{eq:aggr2}
f(\bm{h}(t),\bm{a}(t),t;\bm{\theta}_f) &= f''(\bm{h}''(t),t;\bm{\theta}_f),\\
\bm{h}''(t) &= \bm{h}(t) \odot \phi(\bm{a}(t)),
\end{split}\end{align}\end{linenomath*}where $\odot$ is elementwise multiplication because $\phi(\bm{a}(t))$ contains elementwise attention values. $\phi$ is a sigmoid activation in this case. The other ODE function $g$ also uses $\bm{h}''(t)$.

Note that this definition corresponds to the attention mechanism in image processing, e.g., focusing sub-regions of image for object detection, image classification, and so on. With $\bm{h}''(t)$, we can redefine the ODE function $f(\bm{h}(t),\bm{a}(t),t;\bm{\theta}_f)$ as $f''(\bm{h}''(t),t;\bm{\theta}_f)$.  In our experiments, we also adopt a well-defined ODE function from other work for $f''$. We describe our design for $g$ in the experimental evaluation section since each task or each data has its own optimized attention design.

\subsection{Training Algorithm}
It requires a sophisticated training algorithm to train dual co-evolving NODEs. We describe how to train the proposed model.

It has been reported by several research groups that training NODEs sometimes shows numerical instability, e.g., the underflow error of the adaptive step-size ODE solver. To this end, several advanced regularization concepts (specialized in enhancing the numerical stability of NODEs) have been proposed: i) kinetic energy regularization ii) high-order derivative regularization, and iii) DOPRI's step-size regularization~\cite{kelly2020easynode,finlay2020train}.

\paragraph{Loss Function.} All these regularizations, however, encourage NODEs to learn straight line paths from $t_0$ to $t_1$, which is not adequate for $\bm{a}(t)$. In the architecture in Fig.~\ref{fig:archi}, for instance, we need only $\bm{h}(1)$ for its downstream classification task and therefore, enforcing a straight line for $\bm{h}(t)$ makes sense. However, we cannot say that the trajectory of attention is always straight. The attention value may fluctuate over time and therefore, we consider conventional regularizations such as $L^1$ or $L^2$ regularizations.

Since we adopt existing an NODE design for $\bm{h}(t)$ and extend it to dual co-evolving NODEs by adding $\bm{a}(t)$, we reuse its original setting for $\bm{h}(t)$. We denote its loss function as $L_{\bm{h}}$ --- e.g., $L_{\bm{h}}$ may consist of a take-specific loss, denoted $L_{task}$, and an appropriate regularization term. For our experiments, we do not modify $L_{\bm{h}}$ but strictly follow the original experimental setups to show the efficacy of adding an attention mechanism to it. We then propose the following loss function to train the proposed attention mechanism:
\begin{linenomath*}\begin{align}\begin{split}\label{eq:loss}
    L_{\bm{a}} \stackrel{\text{def}}{=}& L_{task} + \lambda \|\bm{\theta}_g\|_{1\textrm{ or }2},
\end{split}\end{align}\end{linenomath*}where $L_{task}$ is a task-specific loss function, e.g., cross-entropy loss, and $\lambda > 0$ is a coefficient of regularization.

The training algorithm is in Alg.~\eqref{alg:train}. With the above loss functions $L_{\bm{h}}$ and $L_{\bm{a}}$, we alternately train the two NODEs. While training for $\bm{h}(t)$ (resp. $\bm{a}(t)$) with $L_{\bm{h}}$ (resp. $L_{\bm{a}}$), we fix the other NODE $\bm{a}(t)$ (resp. $\bm{h}(t)$). We will describe shortly how to calculate all those gradients except $\nabla_{\bm{\theta}_{others}} L_{\bm{h}}$ in the algorithm. We refer readers to~\cite{NIPS2018_7892} about calculating it.

\begin{algorithm}[t]
\SetAlgoLined
\caption{How to train the dual co-evolving NODEs}\label{alg:train}
\KwIn{Training data $D_{train}$, Validating data $D_{val}$, Maximum iteration number $max\_iter$}
Initialize $\bm{\theta}_f$, $\bm{\theta}_g$, and other parameters $\bm{\theta}_{others}$ if any, e.g., the parameters of the feature extractor and the classifier;

$k \gets 0$;

\While {$k < max\_iter$}{
    Train $\bm{\theta}_f$ and $\bm{\theta}_{others}$ with $\nabla_{\bm{\theta}_f} L_{\bm{h}}$ and $\nabla_{\bm{\theta}_{others}} L_{\bm{h}}$;
    
    Train $\bm{\theta}_g$ with $\nabla_{\bm{\theta}_g} L_{\bm{a}}$;
    
    Validate and update the best parameters, $\bm{\theta}^*_f$, $\bm{\theta}^*_g$ and $\bm{\theta}^*_{others}$, with $D_{val}$\;
    
    $k \gets k + 1$;
}
\Return $\bm{\theta}^*_f$, $\bm{\theta}^*_g$ and $\bm{\theta}^*_{others}$;
\end{algorithm}

\paragraph{Gradient Calculation.} To train them, we need to calculate the gradients of loss w.r.t. NODE parameters $\bm{\theta}_f$ and $\bm{\theta}_g$. In this paragraph, we describe how we can space-efficiently calculate them.

First, we can use the standard adjoint sensitivity method of NODEs to train $\bm{\theta}_f$ because $\bm{\theta}_g$ is fixed and $\bm{a}(t)$ can be considered as constant. Therefore, the gradient of $L_{\bm{h}}$ w.r.t. $\bm{\theta}_f$ can be calculated as follows~\cite{NIPS2018_7892}:
\begin{linenomath*}\begin{align}\begin{split}\label{eq:gra1}
    \nabla_{\bm{\theta}_f} L_{\bm{h}} = \frac{d L_{\bm{h}}}{d \bm{\theta}_f} = -\int_{t_1}^{t_0} \bm{j}_{L_{\bm{h}}}(t)^{\intercal} \frac{\partial f(\bm{h}(t),\bm{a}(t), t;\bm{\theta}_f)}{\partial \bm{\theta}_f} dt,
\end{split}\end{align}\end{linenomath*}where $\bm{j}_{L_{\bm{h}}}(t)$ is an adjoint state defined as $\frac{\partial L_{\bm{h}}}{\partial \bm{h}(t)}$.

Second, it is  more complicated than the previous case to train $\bm{\theta}_g$ because $\bm{a}(t)$ is connected to $L_{\bm{a}}$ via $\bm{h}(t)$. The gradient of $L_{\bm{a}}$ w.r.t $\bm{\theta}_g$ is defined as follows:
\begin{linenomath*}\begin{align}\begin{split}\label{eq:gra2}
    \nabla_{\bm{\theta}_g} L_{\bm{a}} =& \frac{d L_{\bm{a}}}{d \bm{\theta}_g} = \frac{\partial L_{\bm{a}}}{\partial \bm{h}(t)} \cdot \frac{\partial \bm{h}(t)}{\partial \bm{a}(t)} \cdot \frac{d \bm{a}(t)}{d \bm{\theta}_g}\\
    =& -\int_{t_1}^{t_0} \bm{j}_{\bm{h}}(t)^{\intercal} \bm{j}_{L_{\bm{a}}}(t)^{\intercal} \frac{\partial g(\bm{h}(t),\bm{a}(t), t;\bm{\theta}_f)}{\partial \bm{\theta}_g} dt,
\end{split}\end{align}\end{linenomath*}where $\bm{j}_{L_{\bm{a}}}(t)$ is an adjoint state defined as $\frac{\partial L_{\bm{a}}}{\partial \bm{h}(t)}$ and $\bm{j}_{\bm{h}}(t)$ is another adjoint state defined as $\frac{\partial \bm{h}(t)}{\partial \bm{a}(t)}$. Its proof is in Appendix~\ref{a:proof}.

\paragraph{On the Tractability of Training.} The Cauchy–Kowalevski theorem states that, given $f(\bm{h}(t),t;\bm{\theta}_f) = \frac{d \bm{h}(t)}{d t}$, there exists a unique solution of $\bm{h}$ if $f$ is analytic (or locally Lipschitz continuous), i.e., the ODE problem is well-posed when the ODE function is analytic~\cite{10.2307/j.ctvzsmfgn}.

In our case, we alternately train one NODE after fixing the other NODE. When training a NODE, we can assume the mild condition of the analytic ODE function. In our experiments, for instance, we adopt the continuous GRU function from~\cite{debrouwer2019gruodebayes} for $f''$, which consists of analytic operators, e.g., matrix multiplication, matrix addition, hyperbolic tangent, etc. Therefore, the Cauchy–Kowalevski theorem holds for our experiments with continuous GRU cells. In our image classification experiments, however, we use the rectified linear unit (ReLU) activation, which is not analytic. However, our experimental results show that our dual co-evolving NODEs can be trained well.

\section{Experimental Evaluations}
In this section, we describe our experimental environments and results. We conduct experiments with image classification and time-series forecasting. All experiments were conducted in the following software and hardware environments: \textsc{Ubuntu} 18.04 LTS, \textsc{Python} 3.6.6, \textsc{Numpy} 1.18.5, \textsc{Scipy} 1.5, \textsc{Matplotlib} 3.3.1, \textsc{PyTorch} 1.2.0, \textsc{CUDA} 10.0, and \textsc{NVIDIA} Driver 417.22, i9 CPU, and \textsc{NVIDIA RTX Titan}. We repeat the training and testing procedures with five different random seeds and report their mean and standard deviation accuracy. Our source codes are at {\color{blue}\url{https://github.com/sheoyon-jhin/ACE-NODE}}.

\subsection{Image Classification}
\paragraph{Datasets and Baselines.} We use MNIST, SVHN, and CIFAR10, which are all benchmark datasets. We compare our method (ACE-ODE-Net) with ResNet, RKNet, ODE-Net, and Augmented-ODE-Net, following the evaluation protocol used in~\cite{NIPS2018_7892,NIPS2019_8577}. In ResNet, we have a downsampling layer followed by 6 standard residual blocks~\cite{DBLP:conf/eccv/HeZRS16}. For RK-Net and ODE-Net, we replace the residual blocks with a NODE layer but they differ at the choice of ODE solvers. RK-Net uses the fourth-order Runge--Kutta method and ODE-Net uses the adaptive Dormand--Prince method for their forward-pass inference --- both of them are trained with the adjoint sensitivity method which is a standard backward-pass gradient computation method for NODEs. For Augmented-ODE-Net, we follow the augmentation method in~\cite{NIPS2019_8577}. To build our ACE-ODE-Net, therefore, we replace the NODE layer of ODE-Net with our dual co-evolving NODEs without modifying other layers.

\paragraph{Hyperparameters and ODE Functions.} We test the following hyperparameters for our method and other baselines:
\begin{enumerate}
    \item For RKNet and ODE-Net, the ODE function $f$ is shown in Table~\ref{tbl:ode1}. We note that there is no attention in these two models.
    \item For Augmented-ODE-Net, we augment $\bm{h}(t)$ of ODE-Net with five additional dimensions as described in Section~\ref{sec:node}, following the recommendation in~\cite{NIPS2019_8577}.
    \item We do not use the augmentation for ACE-ODE-Net for fair comparison between ACE-ODE-Net and Augmented-ODE-Net.
    \item For the ODE function $f''$ of ACE-ODE-Net, we remove the second layer from Table~\ref{tbl:ode1} to make the model as lightweight as possible.
    \item For the ODE function $g'$ of ACE-ODE-Net, we use the same architecture in Table~\ref{tbl:ode1}.
    \item For the initial attention generator $q$ of ACE-ODE-Net, we adopt similar two convolutional layers.
    \item All in all, our model is the same as ODE-Net except the co-evolving attention.
    \item We train for 160 epochs with a batch size 128 with a learning rate of \{\num{1.0e-3}, \num{5.0e-2}, \num{1.0e-2}\} and a coefficient $\lambda$ of \{\num{1.0e-3}, \num{1.0e-4}\}. While training, we validate with their validation datasets. The default ODE solver is DOPRI.
    \item We note that each dataset has a different image size and the number of parameters for each dataset can vary even with the same overall architecture, i.e., the input and output sizes are different from one dataset to another.
\end{enumerate}

\begin{table}[t]
\centering
\setlength{\tabcolsep}{2pt}
\caption{The architecture of the network $f$ of ODE-Net for MNIST image classification. Conv2D uses a filter size of 3x3, a stride of 1, and a padding of 1.}\label{tbl:ode1}
\begin{tabular}{cccc}
\specialrule{1pt}{1pt}{1pt}
Layer & Design & Input Size & Output Size \\ \specialrule{1pt}{1pt}{1pt}
1 & ReLU(GroupNorm) & $6 \times 6 \times 64$ & $6 \times 6 \times 64$ \\ 
2 & ReLU(GroupNorm(Conv2d)) & $6 \times 6 \times 64$  & $6 \times 6 \times 64$ \\ 
3 & GroupNorm(Conv2d) & $6 \times 6 \times 64$  & $6 \times 6 \times 64$ \\ 
\specialrule{1pt}{1pt}{1pt}
\end{tabular}
\end{table}

\paragraph{Experimental Results.}

\begin{table}[t]
\centering
\setlength{\tabcolsep}{2pt}
\caption{MNIST results}\label{tbl:mnist}
\begin{tabular}{ccc}
\specialrule{1pt}{1pt}{1pt}
Method & Accuracy & \#Params \\ \specialrule{1pt}{1pt}{1pt}
ResNet & 99.65 ± 0.03 & 0.58M \\
RKNet & 98.66 ± 0.57 & 0.21M \\
ODE-Net & 99.61 ± 0.05 & 0.21M \\
Augmented-ODE-Net & 99.66 ± 0.04 & 0.22M \\
ACE-ODE-Net & \textbf{99.68 ± 0.03} & 0.28M \\
\specialrule{1pt}{1pt}{1pt}
\end{tabular}
\end{table}

\begin{table}[t]
\centering
\setlength{\tabcolsep}{2pt}
\caption{SVHN results}\label{tbl:svhn}
\begin{tabular}{ccc}
\specialrule{1pt}{1pt}{1pt}
Method & Accuracy & \#Params \\ \specialrule{1pt}{1pt}{1pt}
ResNet & 95.78 ± 0.05 & 0.58M \\
RKNet & 91.00 ± 0.71 & 0.21M \\
ODE-Net & 95.54 ± 0.13 & 0.21M \\
Augmented-ODE-Net & 95.80 ± 0.08 & 0.22M \\
ACE-ODE-Net & \textbf{96.01 ± 0.10} & 0.32M \\
\specialrule{1pt}{1pt}{1pt}
\end{tabular}
\end{table}

\begin{table}[t]
\centering
\setlength{\tabcolsep}{2pt}
\caption{CIFAR10 results}\label{tbl:cifar}
\begin{tabular}{ccc}
\specialrule{1pt}{1pt}{1pt}
Method & Accuracy & \#Params \\ \specialrule{1pt}{1pt}{1pt}
ResNet & \textbf{86.36 ± 0.41} & 0.58M \\
RKNet & 84.81 ± 0.23 & 0.21M\\
ODE-Net & 85.35 ± 0.62 & 0.21M \\
Augmented-ODE-Net & 85.59 ± 0.09 & 0.22M \\
ACE-ODE-Net & 85.99 ± 0.12 & 0.36M\\
\specialrule{1pt}{1pt}{1pt}
\end{tabular}
\end{table}

\begin{table}
\centering
\setlength{\tabcolsep}{2pt}
\caption{Image Feature Silhouette Score}\label{tbl:cifar_silhouette}
\begin{tabular}{cc}
\specialrule{1pt}{1pt}{1pt}
Method & CIFAR10  \\ \specialrule{1pt}{1pt}{1pt}
ODE-Net & 0.5216 ± 0.003 \\
Augmented-ODE-Net & 0.5220 ± 0.005  \\
ACE-ODE-Net & \textbf{0.5406 ± 0.002}\\
\specialrule{1pt}{1pt}{1pt}
\end{tabular}
\end{table}

\begin{figure}[ht!]
    \centering
    \subfigure[ODE-Net]{\includegraphics[width=0.32\columnwidth]{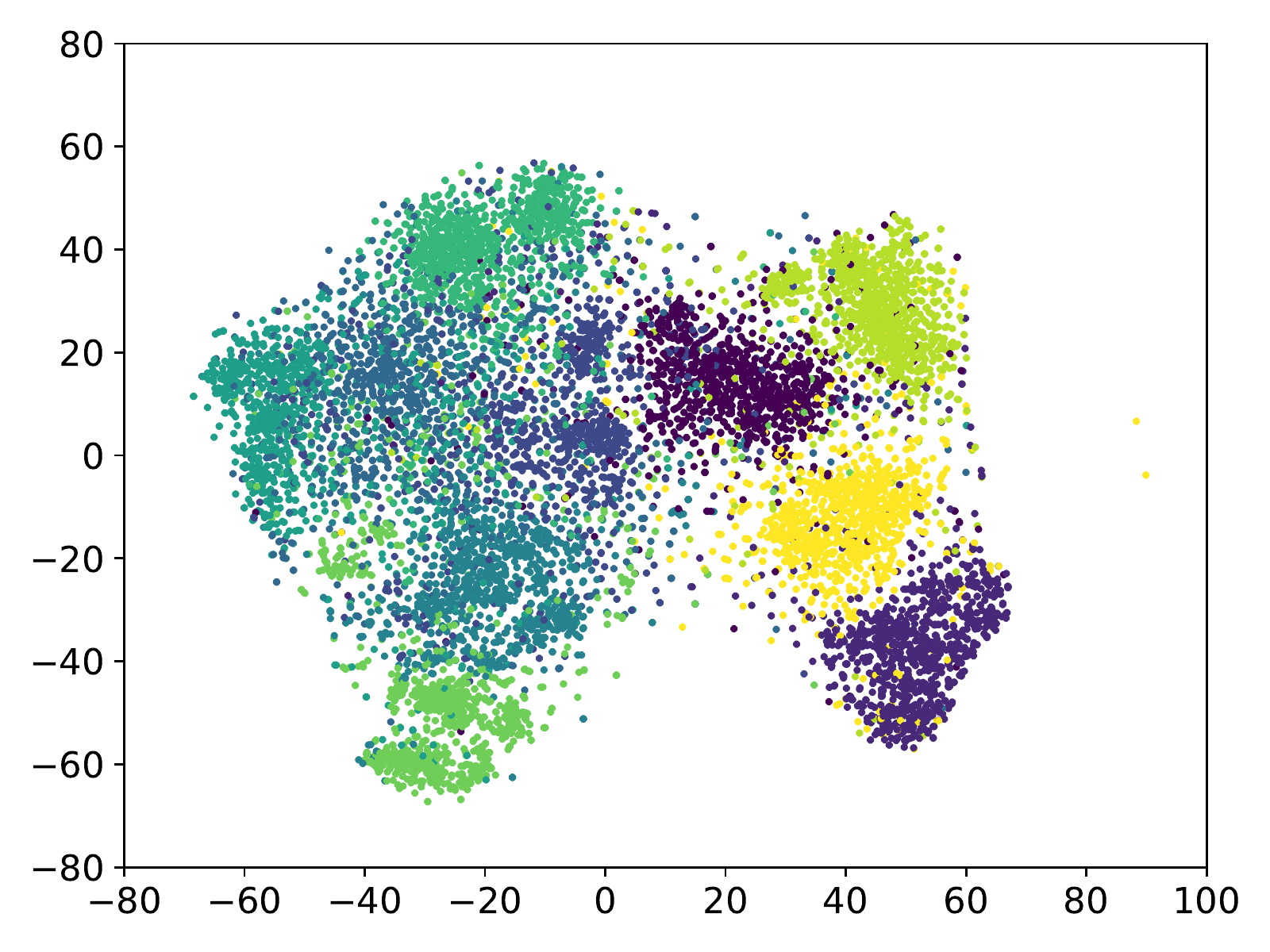}}
    \subfigure[Augmented-ODE-Net]{\includegraphics[width=0.32\columnwidth]{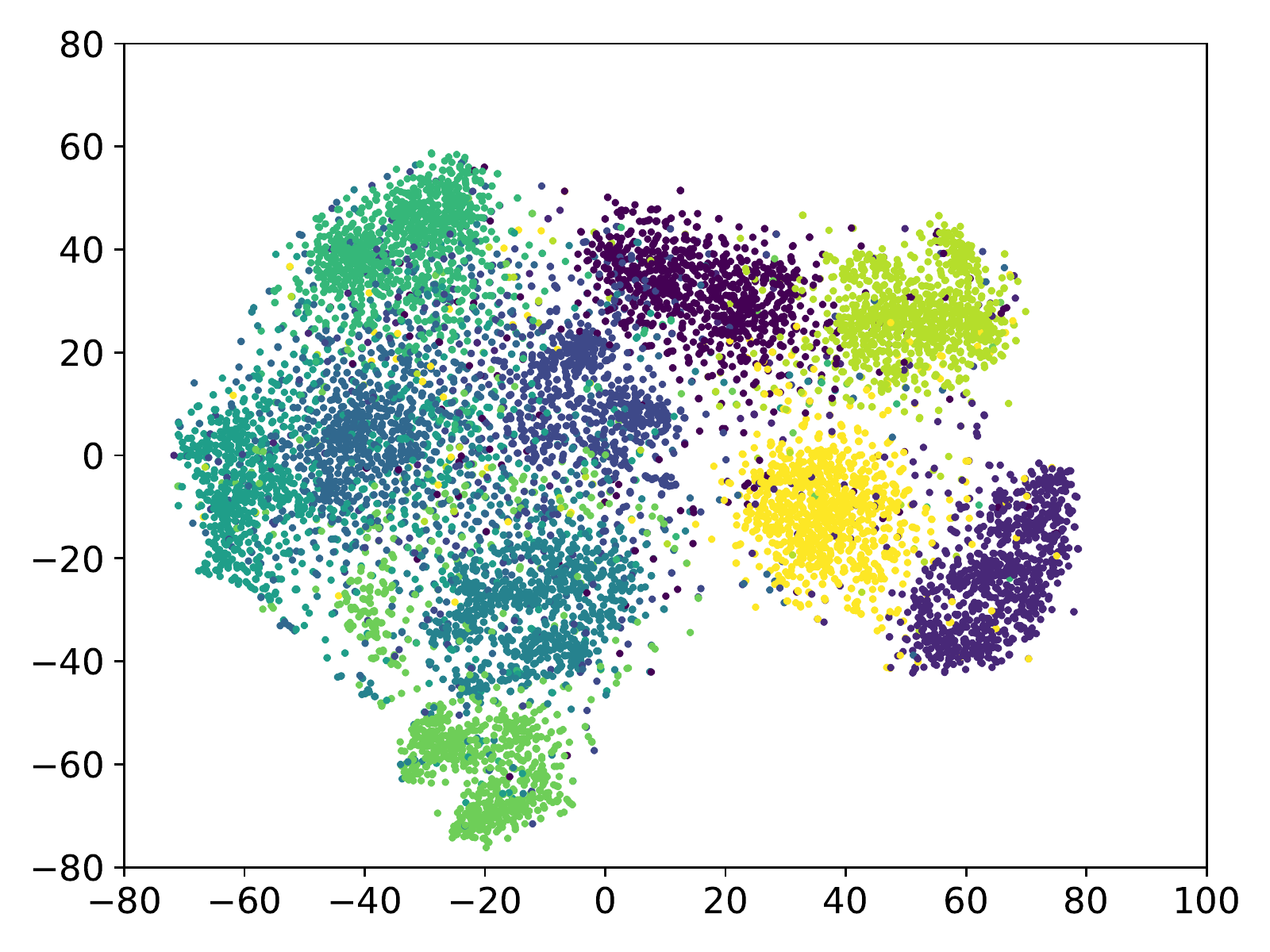}}
    \subfigure[ACE-ODE-Net]{\includegraphics[width=0.32\columnwidth]{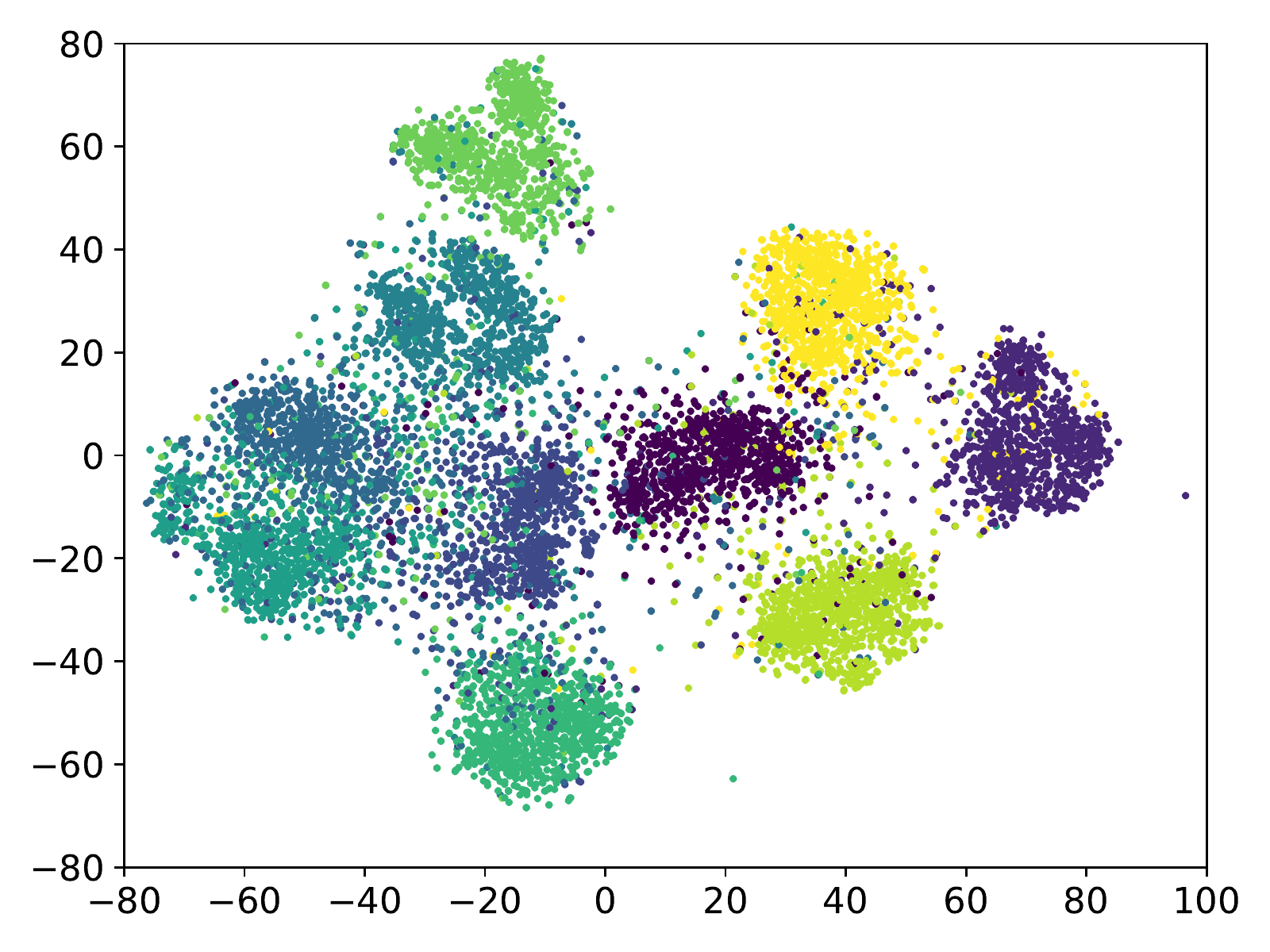}}
    \caption{The visualization of the feature maps for CIFAR10. We use t-SNE to project the feature maps onto a two-dimensional space.}
    \label{fig:tsne2}
\end{figure}

As shown in Table~\ref{tbl:mnist} for MNIST, our ACE-ODE-Net shows the best mean accuracy with the smallest standard deviation. ODE-Net does not outperform ResNet in all cases. Considering that ODE-Net has a smaller number of parameters than that of ResNet, this may be acceptable. Surprisingly, however, our method outperforms ResNet in most cases with a much smaller number of parameters, i.e., 0.58M of ResNet vs. 0.28M of ACE-ODE-Net. This result shows the efficacy of the attention mechanism for NODEs. RKNet shows sometimes unreliable performance than other methods, considering its high standard deviation. In CIFAR10, our method shows the second best accuracy after ResNet. However, ResNet is almost twice as large as our model in terms of the number of parameters. In SVHN, our method shows the best accuracy again.

In order to analyze the efficacy of our attention mechanism in learning image representations, we run the K-Means clustering algorithm with hidden representations generated by various methods and evaluate the quality of clustering them with the silhouette score. As shown in Table~\ref{tbl:cifar_silhouette}, our method shows the largest score, which means the highest quality of clustering. Figure~\ref{fig:tsne2} visualizes the feature maps created by three methods for CIFAR10 and our method shows the best clustering quality in terms of human visual perception. Augmented-ODE-Net slightly improves the clustering quality of ODE-Net. In other datasets, similar patterns are observed.



\subsection{USHCN Climate Forecasting}
\paragraph{Datasets and Baselines.} We reuse the experimental environments of GRU-ODE-Bayes~\cite{debrouwer2019gruodebayes}, a state-of-the-art NODE model for time-series forecasting. We use the U.S. Historical Climatology Network (USHCN) data which contains regional temperature values.

We compare our ACE-GRU-ODE-Bayes with GRU-ODE-Bayes, Augmented-GRU-ODE-Bayes, NODE-VAE, Sequential VAE, and various GRU and LSTM-based models. NODE-VAE uses a two-layer MLP as its ODE function~\cite{NIPS2018_7892}. Sequential-VAE is based on the deep Kalman filter architecture~\cite{krishnan2015deep,10.5555/3298483.3298543}. GRU-Simple, GRU-D, and T-LSTM are all recurrent neural network-based models~\cite{CheEtAl_nature_sr18,10.1145/3097983.3097997}. We extend GRU-ODE-Bayes by replacing its NODE layer with our dual co-evolving NODEs, denoted ACE-GRU-ODE-Bayes.

\paragraph{Hyperparameters and ODE Functions.}
We test the following hyperparameters for our method and other baselines:
\begin{enumerate}
    \item For Augmented-GRU-ODE-Bayes, we augment $\bm{h}(t)$ of GRU-ODE-Bayes with five additional dimensions as described in Section~\ref{sec:node}, following the recommendation in~\cite{NIPS2019_8577}.
    \item We do not use the augmentation for ACE-GRU-ODE-Bayes for fair comparison with Augmented-GRU-ODE-Bayes.
    \item For the ODE functions $f'$ and $g'$ of ACE-GRU-ODE-Bayes, we use the continuous GRU cell proposed in Eq. (3) of~\cite{debrouwer2019gruodebayes}, which is to model GRU cells with ODEs.
    \item For the initial attention generator $q$ of ACE-GRU-ODE-Bayes, we do not use neural networks but calculate a correlation matrix of $\bm{h}(0)$ and set it as $\bm{a}(0)$. This approach results in not only accuracy enhancement but also lightweightness.
    \item All in all, our model is the same as GRU-ODE-Bayes except the co-evolving attention.
    \item We train for 300 epochs with a batch size 300 with a learning rate of \{\num{1.0e-5}, \num{1.0e-4}, \num{1.0e-3}\}, a dropout rate of \{0, 0.1, 0.2, 0.3\}, and a coefficient $\lambda$ of \{0.1, 0.03, 0.01, 0.003, 0.001, 0.0001, 0\}. While training, we use a 5-fold cross validation.
\end{enumerate}

\paragraph{Experimental Results.}

\begin{table}[t]
\centering
\setlength{\tabcolsep}{2pt}
\caption{USHCN-DAILY time-series forecasting results}\label{tbl:ushcn}
\begin{tabular}{ccc}
\specialrule{1pt}{1pt}{1pt}
Method & MSE & NegLL \\ \specialrule{1pt}{1pt}{1pt}
Neural-ODE-VAE & 0.96 ± 0.11 & 1.46 ± 0.10 \\
Neural-ODE-VAE-Mask & 0.83 ± 0.10 & 1.36 ± 0.05 \\
Sequential VAE & 0.83 ± 0.07 & 1.37 ± 0.06 \\
GRU-Simple & 0.75 ± 0.12 & 1.23 ± 0.10 \\
GRU-D & 0.53 ± 0.06 & 0.99 ± 0.07 \\
T-LSTM & 0.59 ± 0.11 & 1.67 ± 0.50 \\
GRU-ODE-Bayes & 0.42 ± 0.04 & 0.98 ± 0.06 \\
Augmented-GRU-ODE-Bayes & 0.44 ± 0.03 & 1.01 ± 0.02 \\
ACE-GRU-ODE-Bayes & \textbf{0.38 ± 0.03} & \textbf{0.94 ± 0.05} \\
\specialrule{1pt}{1pt}{1pt}
\end{tabular}
\end{table}

\begin{figure}
    \centering
    \includegraphics[width=0.8\columnwidth]{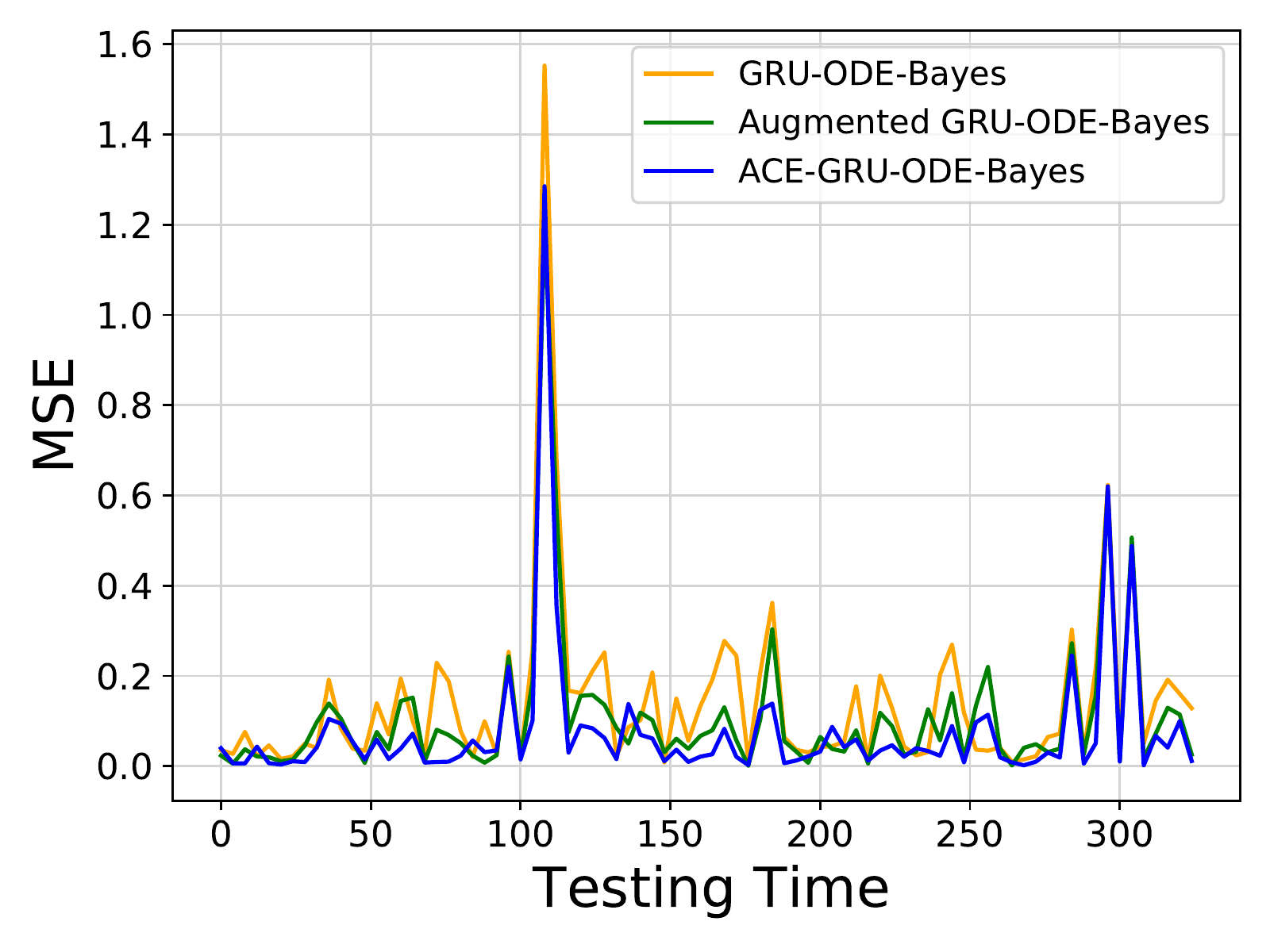}
    \caption{Our method shows smaller errors than others during the almost entire testing time in USHCN.}
    \label{fig:mse}
\end{figure}

We summarize all the results in Table~\ref{tbl:ushcn}. We use mean squared error (MSE) and negative log-likelihood (NegLL) in the table to evaluate each method. Among all baselines, GRU-ODE-Bayes straightly shows the best performance in those two evaluation metrics. Our method outperforms GRU-ODE-Bayes by non-trivial margins, which shows the efficacy of the attention mechanism. In Figure~\ref{fig:mse}, we show the MSE values in the testing period.

\begin{table}[t]
\centering
\setlength{\tabcolsep}{2pt}
\caption{PhysioNet time-series classification results}\label{tbl:physio}
\begin{tabular}{cc}
\specialrule{1pt}{1pt}{1pt}
Method & AUC \\ \specialrule{1pt}{1pt}{1pt}
RNN $\Delta_t$ & 0.787 ± 0.014 \\
RNN-Impute & 0.764 ± 0.016 \\
RNN-Decay & 0.807 ± 0.003 \\
RNN GRU-D & 0.818 ± 0.008 \\
RNN-VAE & 0.515 ± 0.040 \\
Latent-ODE (RNN Enc.) &  0.781 ± 0.018 \\
ODE-RNN & 0.833 ± 0.009 \\
Latent-ODE + Poisson & 0.826 ± 0.007\\
Latent-ODE (ODE Enc.) & 0.829 ± 0.004\\
Augmented-Latent-ODE (ODE Enc.) & 0.851 ± 0.002\\
ACE-Latent-ODE (ODE Enc.) & \textbf{0.853 ± 0.003}\\
\specialrule{1pt}{1pt}{1pt}
\end{tabular}
\end{table}

\subsection{PhysioNet Mortality Classification}
\paragraph{Datasets and Baselines.} The PhysioNet computing in cardiology challenge dataset released at 2012~\cite{doi:10.1177/1460458219850323} is used for this experiment. It is to predict the mortality of intensive care unit (ICU) populations. The dataset had been collected from 12,000 ICU stays. They remove short stays less than 48 hours and recorded up to 42 variables. Each records has a time stamp that indicates an elapsed time after admission to the ICU. Given a record, we predict whether the patient will die or not.

The state-of-the-art baselines for this dataset is various RNN models and NODE models~\cite{NEURIPS2019_42a6845a}. We mainly compare with Latent-ODE which is specialized to irregular time-series datasets.

\paragraph{Hyperparameters and ODE Functions.}
We test the following hyperparameters for our method and other baselines:
\begin{enumerate}
    \item In encoder-decoder baselines, we used 20 latent dimensions in the generative model, 40 dimensions in the recognition model and a batch size of 50. The ODE function of various NODE-based baselines have 3 fully connected (FC) layers with 50 units. We used 20 dimensional hidden state in autoregressive baselines.
    \item For Augmented-Latent-ODE, we augment $\bm{h}(t)$ of Latent-ODE with five additional dimensions as described in Section~\ref{sec:node}, following the recommendation in~\cite{NIPS2019_8577}.
    \item We do not use the augmentation for ACE-Latent-ODE for fair comparison between ACE-Latent-ODE and Augmented-Latent-ODE.
    \item For the ODE functions $f'$ and $g'$ of ACE-Latent-ODE, we use the same 3 fully connected (FC) layers with 50 units.
    \item For the initial attention generator $q$ of ACE-ODE-Net, we do not use neural networks but calculate a correlation matrix of $\bm{h}(0)$ and set it as $\bm{a}(0)$.
    \item All in all, our model is the same as Latent-ODE except the co-evolving attention.
    \item We train for 30 epochs with a batch size 50 with a learning rate of \{\num{1.0e-5}, \num{1.0e-4}, \num{1.0e-3}\}, and a coefficient $\lambda$ of \{0.1, 0.03, 0.01, 0.003, 0.001, 0.0001, 0\}. While training, we validate with their validation datasets.
\end{enumerate}

\paragraph{Experimental Results.} Latent-ODE with our proposed dual co-evolving NODEs, denoted ACE-Latent-ODE, shows the best AUC score. After that, ODE-RNN and Latent-ODE follow. In the original setting, Latent-ODE (ODE Enc.) is inferior to ODE-RNN. After adding our attention, however, it outperforms ODE-RNN, which shows the efficacy of the proposed attention mechanism.

\subsection{PhysioNet Mortality Regression }
\paragraph{Datasets and Baselines.} We use the same PhysioNet dataset for a regression task. In this task, we interpolate and extrapolate the patient vital sign data. Following the work~\cite{NEURIPS2019_42a6845a}, we divide each patient record into a training, a validating, an interpolation test and an extrapolation test periods.

\paragraph{Hyperparameters and ODE Functions.}
We modify the decoder from the encoder-decoder architecture that we have used for the previous classification. The last activation layer of the decoder is changed to a fully connected layer whose output size is 42, i.e., the number of sensors and we solve the integral problem of the NODE-based decoder multiple times according to a set of target time points to predict --- in the previous classification, there is only one sigmoid activation. We also use the same hyperpameter sets to test.

\begin{table}[t]
\centering
\setlength{\tabcolsep}{1pt}
\caption{PhysioNet time-series regression results (MSE)}\label{tbl:physioreg}
\begin{tabular}{ccc}
\specialrule{1pt}{1pt}{1pt}
Method & Interp. (x$10^{-3}$) & Extrap. (x$10^{-3}$) \\ \specialrule{1pt}{1pt}{1pt}
RNN-VAE & 5.930 ± 0.249 & 3.055 ± 0.145 \\
Latent-ODE (RNN enc.) & 3.907 ± 0.252 & 3.162 ± 0.052 \\
Latent-ODE (ODE enc.) & \textbf{2.118 ± 0.271} & 2.231 ± 0.029 \\
Latent-ODE + Poisson & 2.789 ± 0.771 & 2.208 ± 0.050 \\
Augmented-Latent-ODE (ODE enc.) & 2.697 ± 0.771 & 2.042 ± 0.061 \\
ACE-Latent-ODE (ODE enc.) & 2.560 ± 0.08 & \textbf{2.025 ± 0.028} \\
\specialrule{1pt}{1pt}{1pt}
\end{tabular}
\end{table}

\begin{figure}
    \centering
    \includegraphics[width=0.8\columnwidth]{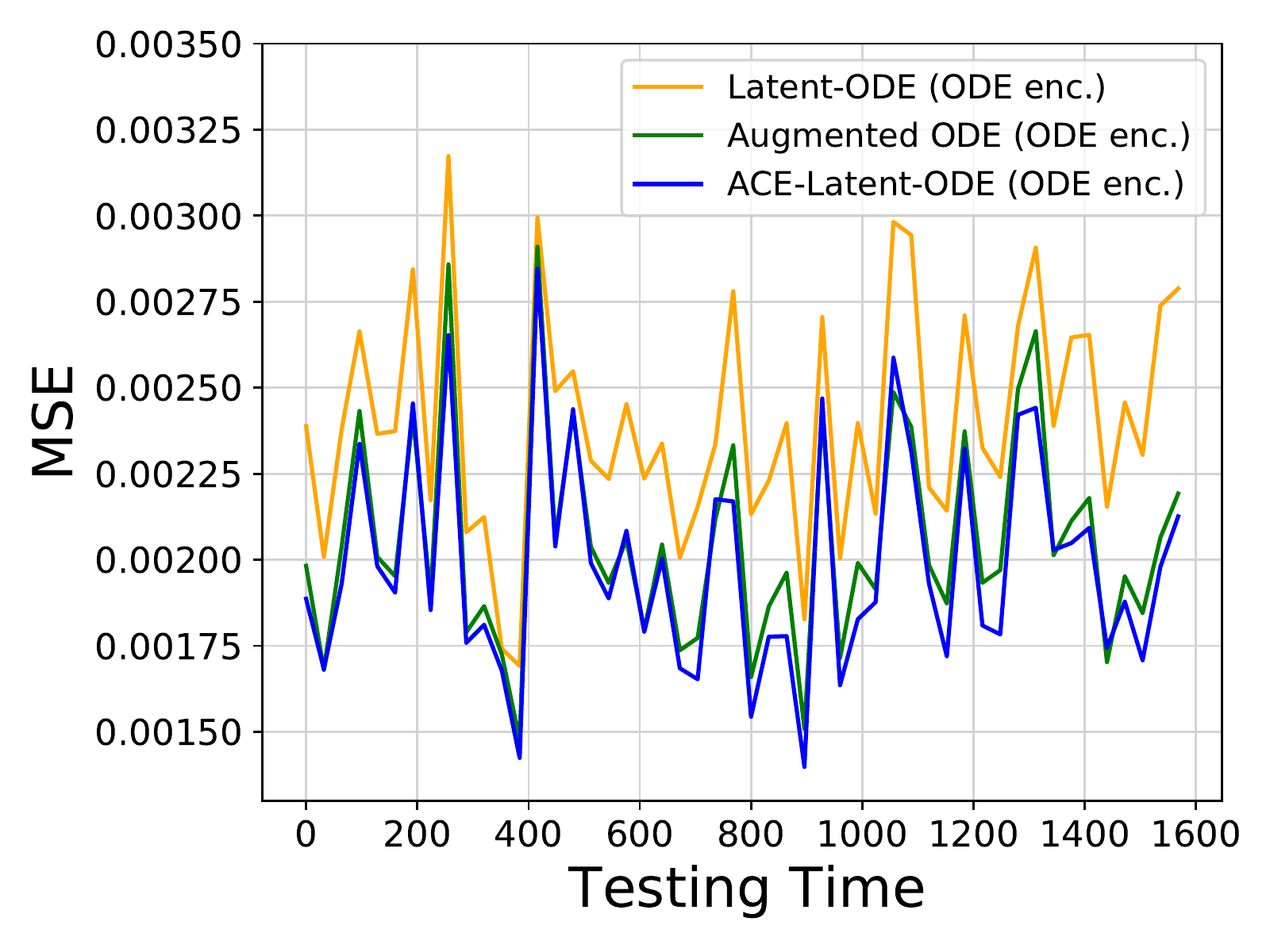}
    \caption{Our method shows smaller errors than others during the almost entire extrapolation testing time in PhysioNet.}
    \label{fig:mse2}
\end{figure}

\paragraph{Experimental Results.} Augmented-Latent-ODE slightly improves Latent-ODE for extrapolation and our ACE-Latent-ODE shows the best accuracy for extrapolation. In Figure~\ref{fig:mse2}, we plot the MSE of various methods during the extrapolation period. For some challenging time points where both Latent-ODE and Augmented-Latent-ODE show relatively higher errors, our ACE-Latent-ODE successfully decreases the errors.

\subsection{Human Activity Classification }

\paragraph{Datasets and Baselines.} The Human Activity dataset~\cite{kaluvza2010activity} contains data from five people performing several activities (walking, falling, lying down, standing up from lying, etc.) with four sensors at left ankle, right ankle, belt and chest. The original data creator let them repeat five times for each performance to collect reliable data. We classify each time point of a person into one of the seven activities in this experiment. The state-of-the-art baseline for this dataset is Latent-ODE~\cite{NEURIPS2019_42a6845a}.

\paragraph{Hyperparameters and ODE Functions.}
We test the following hyperparameters for our method and other baselines:
\begin{enumerate}
    \item In encoder-decoder baselines, we used 15 latent dimensions in the generative model, 100 dimensions in recognition model and batch size of 50. The ODE function of various NODE-based baselines have 3 fully connected (FC) layers with 50 units. We used 15 dimensional hidden state in autoregressive baselines.
    \item For the ODE functions $f''$ and $g$ of ACE-Latent-ODE, we use the same 3 fully connected (FC) layers with 50 units.
    \item For the initial attention generator $q$ of ACE-ODE-Net, we do not use neural networks but calculate a correlation matrix of $\bm{h}(0)$ and set it as $\bm{a}(0)$.
    \item All in all, our model is the same as Latent-ODE except the co-evolving attention.
    \item We train for 70 epochs with a batch size 50 with a learning rate of \{\num{1.0e-5}, \num{1.0e-4}, \num{1.0e-3}\}, and a coefficient $\lambda$ of \{0.1, 0.03, 0.01, 0.003, 0.001, 0.0001, 0\}. While training, we validate with their validation datasets.
\end{enumerate}
\begin{table}[t]
\centering
\setlength{\tabcolsep}{2pt}
\caption{Human Activity time-series classification results}\label{tbl:humanactivity}
\begin{tabular}{cc}
\specialrule{1pt}{1pt}{1pt}
Method & AUC \\ \specialrule{1pt}{1pt}{1pt}
RNN $\Delta_t$ & 0.797 ± 0.003 \\
RNN-Impute & 0.795 ± 0.008 \\
RNN-Decay & 0.800 ± 0.010 \\
RNN GRU-D & 0.806 ± 0.007 \\
RNN-VAE & 0.343 ± 0.040 \\
Latent-ODE (RNN Enc.) &  0.835 ± 0.010 \\
ODE-RNN & 0.829 ± 0.016 \\
Latent-ODE (ODE Enc.) & 0.846 ± 0.013\\
Augmented-ODE (ODE Enc.) & 0.811 ± 0.031\\
ACE-Latent-ODE (ODE Enc.) & \textbf{0.859 ± 0.011}\\
\specialrule{1pt}{1pt}{1pt}
\end{tabular}
\end{table}

\paragraph{Experimental Results.} Our ACE-Latent-ODE significantly outperforms other methods, including Augmented-ODE and Latent-ODE. Surprisingly, Augmented-ODE does not improve Latent-ODE, which sometimes happens. ODE-RNN, which showed good performance for PhysioNet, does not show good accuracy in this dataset.

\subsection{Ablation and Sensitivity Studies}
We conduct additional experiments to show the performance of several variations of our method. Among various ablation and sensitivity study points, we introduce crucial ones in this subsection.

\paragraph{Attention Model Size in Image Classification.} We found that the accuracy sometimes drastically improves when we increase the attention model size, i.e., the number of parameters in $\bm{\theta}_g$. For instance, our ACE-ODE-Net's accuracy improves from 96.01 to 96.12 for SVHN (and improves from 85.99 to 88.12 for CIFAR-10) when we make the number of layers in $g$ large than that of the models in Tables~\ref{tbl:svhn} and~\ref{tbl:cifar}. For MNIST, we do not see any improvements by increasing the capacity of the attention NODE. One possibility is an overfitting in MNIST.

\begin{table}[t]
\centering
\setlength{\tabcolsep}{2pt}
\caption{Image classification results by model size}\label{tbl:imageabl}
\begin{tabular}{cccc}
\specialrule{1pt}{1pt}{1pt}
Method & CIFAR10 & SVHN & MNIST \\ \specialrule{1pt}{1pt}{1pt}
ACE-ODE-Net (Small) & 85.99 \small(0.36M) & 96.01\small(0.32M) & \textbf{99.68}\small(0.28M) \\
ACE-ODE-Net (Large) & \textbf{88.12}\small(1.27M) & \textbf{96.12}\small(1.42M) & 99.58\small(1.27M)\\
\specialrule{1pt}{1pt}{1pt}
\end{tabular}
\end{table}

\begin{figure}
    \centering
    \includegraphics[width=0.8\columnwidth]{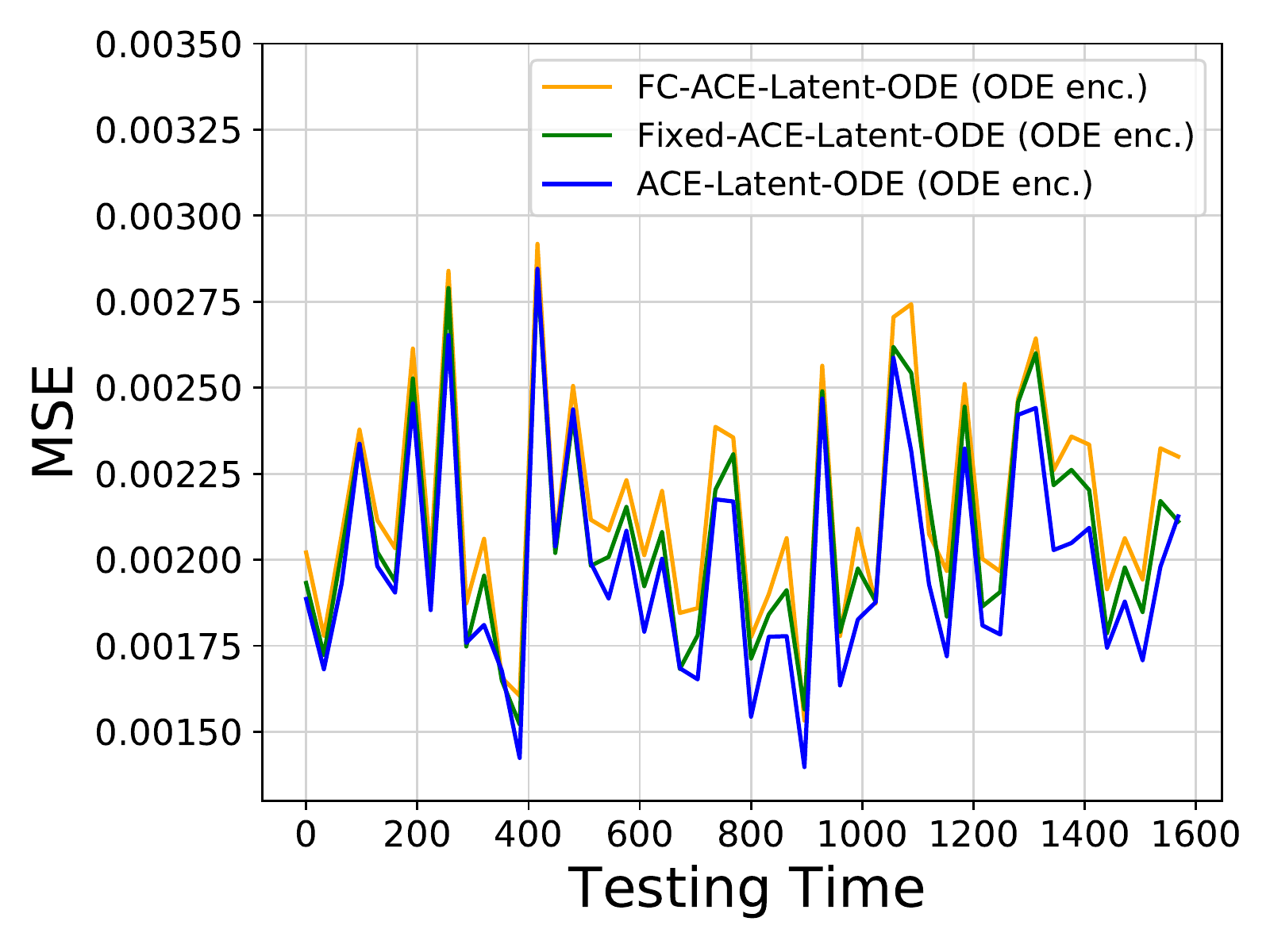}
    \caption{Our attention shows smaller errors than other FC-based and fixed attention methods in PhysioNet.}
    \label{fig:mse3}
\end{figure}

\paragraph{Initial Attention for Time-series Experiments.} For time-series datasets, we found that it is important how to initialize $\bm{a}(0)$. Surprisingly, we found that using a correlation matrix of $\bm{h}(0)$ is better than using other options, such as FC layers, etc. When we use FC layers, denoted as FC-ACE-ODE-Net in Table~\ref{tbl:imageabl} and Figure~\ref{fig:mse3}, our method's MSE increases from 0.38 to 0.55 for USHCN-DAILY and AUC decreases from 0.853 to 0.793 for PhysioNet. In PhysioNet, our method's MSE increases from 2.025(x$10^{-3}$) to 2.108(x$10^{-3}$).

\paragraph{Fixed Attention for Time-series Experiments.} We also let $\bm{a}(t)$ be a correlation matrix of $\bm{h}(t)$, for all $t$, instead of evolving it from $\bm{a}(0)$ through a NODE, denoted as Fixed-ACE-ODE-Net in Table~\ref{tbl:imageabl} and Figure~\ref{fig:mse3}. In this case, the AUC score of our method decreases from 0.853 to 0.838 for PhysioNet. For GRU-ODE-Bayes, it outputs nonsensical values. In PhysioNet, our method's MSE increases from 2.025(x$10^{-3}$) to 2.113(x$10^{-3}$). Figure~\ref{fig:mse3} shows their errors over testing time.

\begin{table}[t]
\centering
\setlength{\tabcolsep}{2pt}
\caption{Time-series classification results by attention type}\label{tbl:imageabl}
\begin{tabular}{ccc}
\specialrule{1pt}{1pt}{1pt}
Method & PhysioNet\small(AUC) & Human Activity\small(AUC) \\ \specialrule{1pt}{1pt}{1pt}
FC-ACE-ODE-Net & 0.793 & 0.569 \\
Fixed-ACE-ODE-Net& 0.838 & 0.551 \\
ACE-ODE-Net& \textbf{0.853} & \textbf{0.859} \\
\specialrule{1pt}{1pt}{1pt}
\end{tabular}
\end{table}

\section{Conclusions}
Attention is widely used for various machine learning tasks. However, it has been overlooked for a while for NODEs. We presented a method of dual co-evolving NODEs to describe cross-evolutionary processes of hidden vector and attention.

To show the efficacy of the method, we conducted in-depth experiments with various state-of-the-art NODE models for various downstream tasks, ranging from image classification to time-series forecasting. Our method consistently outperforms existing NODE-based baselines. We also found that it is important how to initialize the initial attention, denoted $\bm{a}(0)$, in the pairwise attention and proposed to use a correlation matrix of $\bm{h}(0)$. Our ablation study showed that it is a better choice than other neural network-based initializations.


\begin{acks}
Noseong Park is the corresponding author. This work was supported by the Institute of Information \& Communications Technology Planning \& Evaluation (IITP) grant funded by the Korea government (MSIT) (No. 2020-0-01361, Artificial Intelligence Graduate School Program (Yonsei University)).
\end{acks}

\bibliographystyle{ACM-Reference-Format}
\bibliography{ijcai21}

\clearpage
\appendix
\section{Proof}\label{a:proof}
\begin{theorem}
The gradient of the loss $L_{\bm{a}}$ w.r.t $\bm{\theta}_g$ is defined as follows:
\begin{linenomath*}\begin{align*}\begin{split}
    \nabla_{\bm{\theta}_g} L_{\bm{a}} = -\int_{t_1}^{t_0} \bm{j}_{\bm{h}}(t)^{\intercal} \bm{j}_{L_{\bm{a}}}(t)^{\intercal} \frac{\partial g(\bm{h}(t),\bm{a}(t), t;\bm{\theta}_f)}{\partial \bm{\theta}_g} dt.
\end{split}\end{align*}\end{linenomath*}
\end{theorem}
\begin{proof}
Let us define an adjoint state $\bm{j}(t)$ as $\frac{\partial L_{\bm{a}}}{\partial \bm{a}(t)}$. According to the standard adjoint sensitivity proof~\cite{NIPS2018_7892},
\begin{linenomath*}\begin{align*}\begin{split}
    \nabla_{\bm{\theta}_g} L_{\bm{a}} = -\int_{t_1}^{t_0} \bm{j}(t)^{\intercal} \frac{\partial g(\bm{h}(t),\bm{a}(t), t;\bm{\theta}_f)}{\partial \bm{\theta}_g} dt.
\end{split}\end{align*}\end{linenomath*}

However, the adjoint state $\bm{j}(t)$ can be further derived to $\bm{j}_{L_{\bm{a}}}(t) \bm{j}_{\bm{h}}(t)$, where $\bm{j}_{L_{\bm{a}}}(t) = \frac{\partial L_{\bm{a}}}{\partial \bm{h}(t)}$ and $\bm{j}_{\bm{h}}(t) = \frac{\partial \bm{h}(t)}{\partial \bm{a}(t)}$.

After substituting $\bm{j}(t)$ with $\bm{j}_{L_{\bm{a}}}(t) \bm{j}_{\bm{h}}(t)$,
\begin{linenomath*}\begin{align*}\begin{split}
    \nabla_{\bm{\theta}_g} L_{\bm{a}} =& -\int_{t_1}^{t_0} \big(\bm{j}_{L_{\bm{a}}}(t) \bm{j}_{\bm{h}}(t)\big)^{\intercal} \frac{\partial g(\bm{h}(t),\bm{a}(t), t;\bm{\theta}_f)}{\partial \bm{\theta}_g} dt,\\
    =& -\int_{t_1}^{t_0} \bm{j}_{\bm{h}}(t)^{\intercal} \bm{j}_{L_{\bm{a}}}(t)^{\intercal} \frac{\partial g(\bm{h}(t),\bm{a}(t), t;\bm{\theta}_f)}{\partial \bm{\theta}_g} dt.
\end{split}\end{align*}\end{linenomath*}
\end{proof}

\section{Best Hyperparameters}
We list the best hyperparameter with which we produced all the reported results as follows:
\begin{enumerate}
    \item For MNIST, $\lambda$=\num{1.0e-4}, learning rate = \num{1.0e-2}, Relative Tolerance = \num{1.0e-4}, Absolute Tolerance = \num{1.0e-4};
    \item For SVHN, $\lambda$=\num{1.0e-3}, learning rate = \num{5.0e-2},Relative Tolerance = \num{1.0e-5}, Absolute Tolerance = \num{1.0e-5};
    \item For CIFAR10, $\lambda$=\num{1.0e-3}, learning rate = \num{1.0e-2},Relative Tolerance = \num{1.0e-5}, Absolute Tolerance = \num{1.0e-5};
    \item For USHCN-DAILY, $\lambda$=\num{1.0e-2}, learning rate = \num{1.0e-4}, dropout rate = 0.2, Weight Decay = \num{1.0e-4};
    \item For PhysioNet, $\lambda$=\num{1.0e-2}, learning rate = \num{1.0e-2}, Relative Tolerance = \num{1.0e-4}, Absolute Tolerance = \num{1.0e-5};.
    \item For Human-Activity, $\lambda$=\num{1.0e-2}, learning rate = \num{1.0e-2}, Relative Tolerance = \num{1.0e-4}, Absolute Tolerance = \num{1.0e-5}.
\end{enumerate}

\section{Architecture of the network f of ODE-NET for Experiments}
We show the architecture of the ODE functions we used for our experiments in Tables~\ref{tbl:ode3} to \ref{tbl:ode5}.

\begin{table}[H]
\centering
\setlength{\tabcolsep}{2pt}
\caption{The architecture of the network $f$ of ODE-Net for CIFAR10 image classification Task. Conv2D uses a filter size of 3x3, a stride of 1, and a padding of 1.}\label{tbl:ode3}
\begin{tabular}{cccc}
\specialrule{1pt}{1pt}{1pt}
Layer & Design & Input Size & Output Size \\ \specialrule{1pt}{1pt}{1pt}
1 & ReLU(GroupNorm) & $7 \times 7 \times 64$ & $7 \times 7 \times 64$ \\ 
2 & ReLU(GroupNorm(Conv2d)) & $7 \times 7 \times 64$  & $7 \times 7 \times 64$ \\ 
3 & GroupNorm(Conv2d) & $7 \times 7 \times 64$  & $7 \times 7 \times 64$ \\ 
\specialrule{1pt}{1pt}{1pt}
\end{tabular}
\end{table}

\begin{table}[H]
\centering
\setlength{\tabcolsep}{2pt}
\caption{The architecture of the network $f$ of ODE-Net for SVHN image classification Task. Conv2D uses a filter size of 3x3, a stride of 1, and a padding of 1.}\label{tbl:ode3}
\begin{tabular}{cccc}
\specialrule{1pt}{1pt}{1pt}
Layer & Design & Input Size & Output Size \\ \specialrule{1pt}{1pt}{1pt}
1 & ReLU(GroupNorm) & $8 \times 8 \times 64$ & $8 \times 8 \times 64$ \\ 
2 & ReLU(GroupNorm(Conv2d)) & $8 \times 8 \times 64$  & $8 \times 8 \times 64$ \\ 
3 & GroupNorm(Conv2d) & $8 \times 8 \times 64$  & $8 \times 8 \times 64$ \\ 
\specialrule{1pt}{1pt}{1pt}
\end{tabular}
\end{table}

\begin{table}[H]
\centering
\setlength{\tabcolsep}{2pt}
\caption{The architecture of the network $f$ of Latent-ODE for PhysioNet (classification, regression)}.\label{tbl:ode4}
\begin{tabular}{cccc}
\specialrule{1pt}{1pt}{1pt}
Layer & Design & Input Size & Output Size \\ \specialrule{1pt}{1pt}{1pt}
1 & Tanh(Linear) & $3 \times 50 \times 20$ & $3 \times 50 \times 50$ \\ 
2 & Tanh(Linear) & $3 \times 50 \times 50$ & $3 \times 50 \times 50$ \\ 
3 & Tanh(Linear) & $3 \times 50 \times 50$ & $3 \times 50 \times 50$ \\ 
4 & Tanh(Linear) & $3 \times 50 \times 50$ & $3 \times 50 \times 20$ \\ 
\specialrule{1pt}{1pt}{1pt}
\end{tabular}
\end{table}

\begin{table}[H]
\centering
\setlength{\tabcolsep}{2pt}
\caption{The architecture of the network $f$ of Latent-ODE for Human Activity (classification)}.\label{tbl:ode5}
\begin{tabular}{cccc}
\specialrule{1pt}{1pt}{1pt}
Layer & Design & Input Size & Output Size \\ \specialrule{1pt}{1pt}{1pt}
1 & Tanh(Linear) & $3 \times 50 \times 15$ & $3 \times 50 \times 500$ \\ 
2 & Tanh(Linear) & $3 \times 50 \times 500$ & $3 \times 50 \times 500$ \\ 
3 & Tanh(Linear) & $3 \times 50 \times 500$ & $3 \times 50 \times 500$ \\ 
4 & Linear & $3 \times 50 \times 500$ & $3 \times 50 \times 15$ \\ 
\specialrule{1pt}{1pt}{1pt}
\end{tabular}
\end{table}

\end{document}